\documentclass[letterpaper, 10 pt, conference]{ieeeconf}  

\IEEEoverridecommandlockouts                              

\usepackage{lipsum,adjustbox,makecell,multirow,bm,capt-of,etoolbox, amsmath, authblk, booktabs, hyperref}

\begin{document}

\title{\LARGE \bf
WaveShot: A Compact Portable Unmanned Surface Vessel for Dynamic Water Surface Videography and Media Production
}

\author{
Shijian Ma$^{1*}$, Shicong Ma$^{2*}$, Jianhao Jiao$^{3}$
\thanks{$^{*}$Equal contribution.}
\thanks{$^{1}$University of Macau, Avenida da Universidade
Taipa, Macau, China. mc36473@um.edu.mo}
\thanks{$^{2}$Independent researcher. alexma3312@gmail.com}
\thanks{$^{3}$Department of Computer Science, University College London, Gower Street, WC1E 6BT, London, UK. ucacjji@ucl.ac.uk}
\thanks{Special thanks to Greenbay Autopilot Technology Co., Ltd. for their technical support in unmanned vessel hardware.}
}

\maketitle

\begin{abstract}

This paper presents WaveShot, an innovative portable unmanned surface vessel that aims to transform water surface videography by offering a highly maneuverable, cost-effective, and safe alternative to traditional filming methods. WaveShot is specially designed for the modern demands of film production, advertising, documentaries, and visual arts, equipped with professional-grade waterproof cameras and advanced technology to capture both static and dynamic scenes on waterways. We discuss the development and advantages of WaveShot, highlighting its portability, ease of transport, and rapid deployment capabilities. Experimental validation that is showcasing WaveShot's stability and high-quality video capture in various water conditions, and the integration of monocular depth estimation algorithms to enhance the operator's spatial perception. The paper concludes with an exploration of WaveShot's real-world applications, its user-friendly remote operation, and future enhancements such as gimbal integration and advanced computer vision for optimized videography on water surfaces.

\end{abstract}


\begin{figure*}[t]
    \centering
    \includegraphics[width=\textwidth]{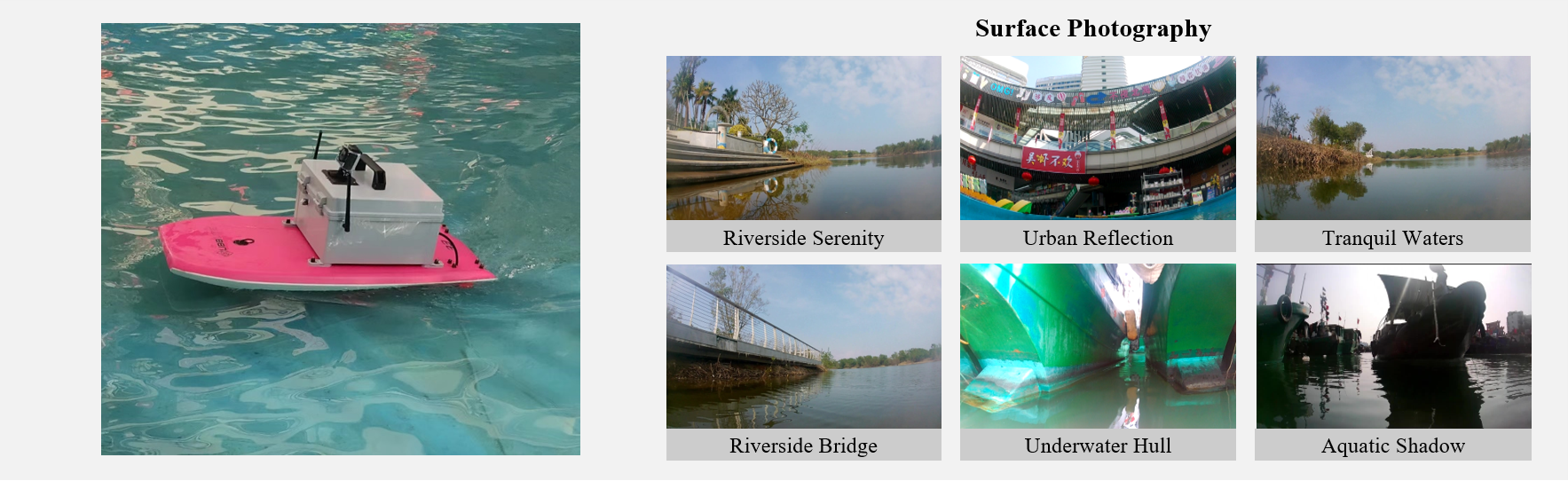}
    \caption{\textbf{\textit{WaveShot}}. We introduce a compact portable unmanned surface vessel for dynamic water surface videography and media production. \textit{Left:} WaveShot is sailing in the pool with a stable posture and recording videos of the water surface. \textit{Right:} WaveShot can perform water surface filming tasks in various water environments. For a live demonstration, see the \href{https://youtu.be/5thT7o9Us5Q}{demo video}.}
    \label{fig:teaser}
\end{figure*}

\section{INTRODUCTION}

In recent years, Unmanned Surface Vehicles (USVs) have attracted increasing research interest due to their potential advantages in efficiency, cost, and safety. Extensive work has been done in the development of USV platforms and core technologies. Many studies have focused on designing USV platforms for various applications. Bibuli et al. \cite{Bibuli2008LineFG} developed an autonomous catamaran named Charlie for environmental monitoring tasks. Campbell et al. \cite{CAMPBELL2012267} proposed a USV named Measuring Dolphin for bathymetric measurements. A low-cost USV named WeMo, which uses a LoRa wireless network for real-time long-distance water quality monitoring tasks, was developed by Madeo et al. \cite{8948249}. Research on unmanned solar USVs as launch and recovery platforms for Unmanned Aerial Vehicles (UAVs) includes an innovative concept design proposed by Aissi et al. \cite{9314415}, achieving autonomous launching and recovery of UAVs as well as automatic battery charging in unmanned mode, providing a viable solution for long-term maritime operations.

Some research has been dedicated to the navigation, guidance, and control methods of USVs. Do et al. \cite{DO2004929} modeled the motion dynamics of USVs and developed a path-following controller. Lee et al. \cite{Lee2004AFL} introduced a fuzzy logic-based method for USV navigation and collision avoidance. Wei et al. \cite{10280968} presented a local obstacle avoidance method for USVs based on the Vector Field Histogram. Liu et al. \cite{LIU20181550} utilized variable space-time optimization to conform USV paths to maritime rules.

Despite extensive research and development in the field of USVs, there is a noticeable gap in their application in professional waterborne filming tasks. Current USV platforms primarily focus on environmental monitoring, measurement, and security operations, with less attention given to applications using action cameras for professional filming purposes. To bridge this gap, we introduce WaveShot, a compact and portable USV specifically designed for capturing high-quality video material on water using waterproof action cameras. WaveShot aims to leverage advancements in USV technology to provide a unique platform for capturing high-quality video footage in aquatic environments, marking a new application of USVs in film photography and media production.

In addition to hardware innovation, integrating Monocular Depth Estimation (MDE) technology into WaveShot's framework adds depth and detail to the footage. Initially characterized by manually designed features and probabilistic graphical models for depth inference, Saxena et al. \cite{NIPS2005_17d8da81} adopted a supervised learning approach, using Markov Random Fields (MRF) to predict depth information from a single image, training their model based on local and global features extracted from the image. With the advent of deep learning technologies, Convolutional Neural Networks (CNNs) have become the core technology for monocular depth estimation. Eigen et al. \cite{eigen2014depth} developed a multi-scale CNN framework that improved depth prediction by learning from coarse to fine, achieving better results on standard datasets compared to traditional techniques.

To improve the accuracy and efficiency of monocular depth estimation, researchers have been actively exploring the application of deep learning. For example, Laina et al. \cite{laina2016deeper} designed an advanced Fully Convolutional Residual Network (FCRN) that performed excellently across various scenes. They also introduced a novel reverse Huber loss function that helped to adapt to a wide range of depth variations and promote smooth depth transitions. Unsupervised learning methods have also garnered interest as they do not require real depth labels for training. Garg et al. \cite{garg2016unsupervised} showcased an unsupervised learning framework that trained CNNs to predict depth by minimizing the photometric reconstruction error of stereo image pairs. Their method was strengthened by combining left-right consistency constraints \cite{godard2017unsupervised} with temporal consistency constraints \cite{zhou2017unsupervised}. To enhance the performance of monocular depth estimation across different scenes, researchers have also investigated transfer learning and domain adaptation techniques. Godard et al. \cite{godard2019digging} proposed a self-supervised learning strategy, training models on diverse datasets so they could accurately predict the depth of any given image. The MiDaS \cite{ranftl2020robust} project introduced an affine-invariant loss function, effectively mitigating depth scale and offset differences between datasets and achieving efficient cross-dataset joint training. In this paper, we employ the Depth Anything \cite{yang2024depth} zero-shot depth estimation model from MDE technology, combined with video captured by WaveShot on water surfaces, to generate sequences of consecutive depth estimation images.

In this paper, we design and implement WaveShot in Figure~\ref{fig:teaser}, a portable USV platform for waterborne target filming. Our main contributions include: 1) Improving the flexibility and convenience of the filming process by optimizing the size and weight of WaveShot, while ensuring that the quality of filming is not compromised. 2) We validate the effectiveness of WaveShot in filming both static and moving water scenes through a series of experiments. The experimental results show that WaveShot can provide stable, clear video capture under various water conditions. Applying the latest monocular depth estimation algorithm adds depth information to the filmed waterborne targets, offering new perspectives.

The rest of the paper is organized as follows. We describe the details of WaveShot's hardware architecture in Section II. In Section III, we introduce the related tasks of WaveShot, and in Section IV, we validate WaveShot through experiments. Section V concludes the paper.


\begin{figure*}[t!]
    \centering
    \begin{minipage}[c]{0.7\textwidth}
        \centering
        \includegraphics[width=\linewidth]{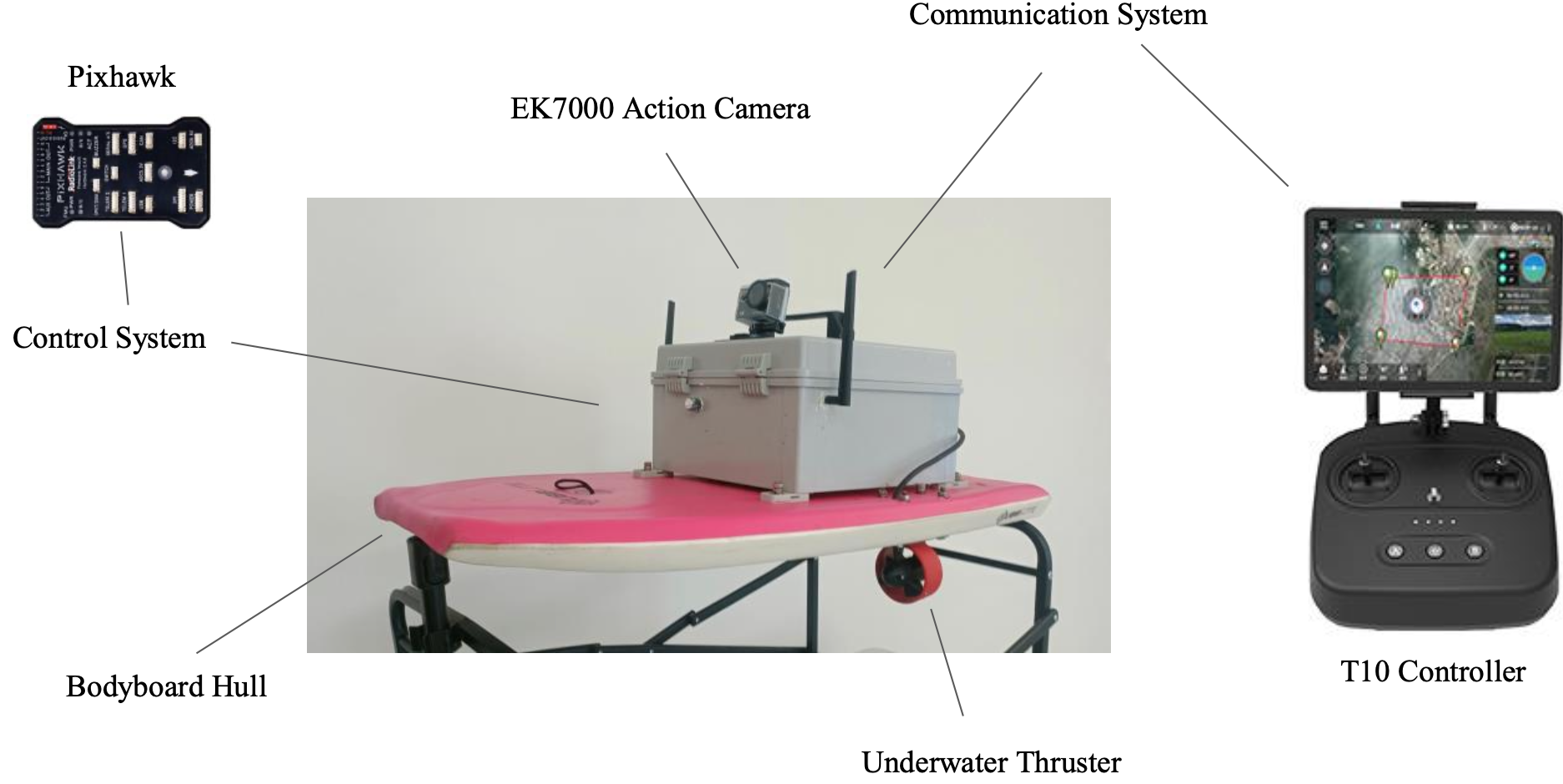} 
    \end{minipage}%
    \begin{minipage}[c]{0.3\textwidth}
        \centering
        \includegraphics[width=\linewidth]{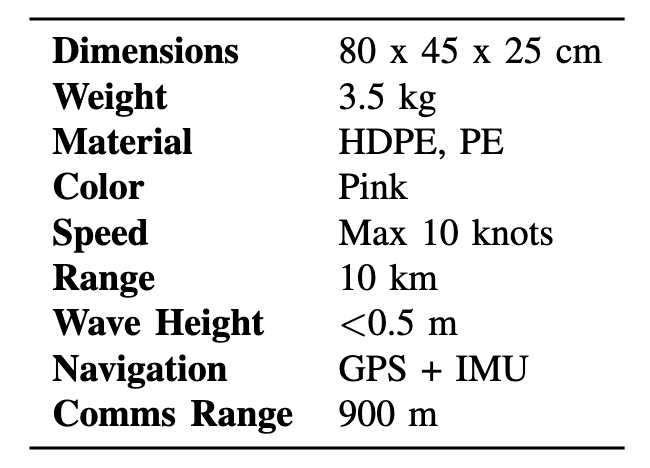} 
    \end{minipage}
    \caption{\textbf{\textit{Hardware Details.}} \textit{Left:} WaveShot consists of a bodyboard hull, control system, communication system, underwater thruster, and an action camera on the head. \textit{Middle:} Manual operation is set up using the SKYDROID T10 Controller combined with a mobile ground station, allowing for precise navigation of WaveShot within a range of 900 meters. \textit{Right:} Technical specifications of WaveShot.}
    \label{fig:hardware_details}
\end{figure*}

\section{WAVESHOT HARDWARE}

We have developed WaveShot, a portable waterborne filming boat, for capturing footage and executing filming tasks in various aquatic environments. WaveShot features low cost, flexibility, and simple automatic filming capabilities, integrating various AI visual technologies into the filmed videos. Specifically, we have incorporated four key design considerations:

\begin{enumerate}
\item \textbf{Autonomous Navigation}: The system is capable of autonomous navigation, facilitating hands-free operation in diverse water environments.
\item \textbf{Stability}: It maintains stability even under windy and wavy conditions.
\item \textbf{Full-Body Remote Operation}: The boat body and camera, can be remotely controlled.
\item \textbf{Portability}: The design is ensuring easy transportation and quick deployment.
\end{enumerate}

As illustrated in Figure~\ref{fig:hardware_details}, we chose a bodyboard as the hull for the boat. A bodyboard is a wave-riding board shorter than a surfboard, typically around 0.8 meter in length. Made of foam, it provides good buoyancy and stability on the water. Designed to support a person floating on the water, it ensures even distribution of buoyancy across its surface, maintaining stability even in complex water conditions or when carrying heavy loads. Lightweight materials such as high-density polyethylene (HDPE) and polyethylene (PE) provide buoyancy without adding excessive weight. The shape and contour of the bodyboard are designed to effectively cut through water, enhancing stability and maneuverability. We found that the bodyboard can withstand prolonged exposure to seawater and sunlight, which is advantageous for unmanned vessels operating in harsh environments. It is very lightweight, aiding in transportation and maneuverability, making efficient propulsion systems possible. Modifying and installing additional equipment on the hull is straightforward.

Next, we designed a waterproof control system on the bodyboard, capable of simultaneously controlling two external underwater thrusters and multiple sensors. This design choice is crucial for ensuring WaveShot's stability in water environments along predefined routes during filming. We installed a flip-top waterproof box on the boat's body to house the control board and sensor components. The selected control board is the open-source Pixhawk hardware, containing a high-performance microprocessor and sensors, including gyros, accelerometers, magnetometers, and barometers, ensuring precise navigation control and stability. We flashed ArduPilot firmware \cite{ardupilot} to support autonomous navigation of differential boats. To achieve autonomous navigation in various bodies of water, we integrated a GPS module onto Pixhawk, providing accurate geographic location information critical for route planning and heading control.

WaveShot utilizes the precision of PID control algorithms to achieve high performance in path following, with each control aspect meticulously tuned within the ArduPilot framework \cite{ardupilot}. The PID controller's output, \( u(t) \), dictates the necessary adjustments to minimize the error signal \( e(t) \), which is the deviation from the desired trajectory. It is defined as follows:

\[
u(t) = K_p \cdot e(t) + K_i \int_{0}^{t} e(\tau) d\tau + K_d \frac{d}{dt} e(t)
\]

Within this context, \( K_p \), \( K_i \), and \( K_d \) are set to optimize the board's response. The \texttt{ATC\_STR\_ACC\_MAX} parameter is set to \( 120 \text{ cm/s}^2 \), moderating the maximum achievable acceleration to prevent thruster saturation. \texttt{ATC\_STR\_ANG\_P} is assigned a value of \( 1 \), tuning the proportional response to the current heading error. \texttt{ATC\_STR\_RAT\_D} with a value of \( 0.02 \) tempers the rate of change of the error, providing stability. \texttt{ATC\_STR\_RAT\_MAX} is set to \( 30^\circ/\text{s} \), limiting the maximum steering rate to prevent overly aggressive maneuvers. Lastly, \texttt{ATC\_STR\_RAT\_I} at \( 0.2 \) ensures that any persistent offset from the desired course is corrected by integrating the error over time. These parameter values collectively contribute to WaveShot's superior and precise path-following performance.

The communication system in the WaveShot control system ensures a stable connection between the boat and the ground station, allowing for transmission of control commands and reception of boat status information. This communication system includes radio transmission devices, antennas, and ground stations. We externally mounted two antennas on the control box of the boat to enhance signal reception and transmission, with a stable communication range of up to one kilometer during water tests. We used Mission Planner ground station software on the computer and QGroundControl ground station software on the phone to display real-time data of the boat, map positioning, flight planning, and monitor boat status, adjusting task parameters as needed. Additionally, we installed a receiver paired with a remote controller for manual operation.

Its power system is the core part of maneuvering and controlling the boat. We mounted two underwater thrusters on the bottom of the boat, using two underwater brushless thrusters as power sources, a configuration known as a differential propulsion system. The differential propulsion system achieves the boat's forward, backward, turning, and stopping movements by independently controlling the speeds of the two thrusters, providing good maneuverability and control accuracy. By simultaneously increasing or decreasing the speeds of the two thrusters, the boat can move forward or backward. The magnitude of the speed determines the boat's velocity. By changing the speed difference between the two thrusters, the boat can turn left or right. For example, increasing the speed of the right thruster and decreasing the speed of the left thruster will turn the boat to the left. The brushless thrusters operate fully submerged in water, with heat generated during operation dissipating through water cooling.

Above the control box, we installed an EK7000 action camera with a waterproof housing, capable of recording ultra-high-definition 4K videos and waterproof functionality in Figure~\ref{fig:hardware_details}. It also comes with a remote control and WiFi functionality for remote wireless video recording start-up and transmission, allowing us to collect video data accurately on demand and share it with others. Adjusting WaveShot's perspective and trajectory based on video feedback to improve filming quality. Some example tasks that WaveShot combined with computer vision technology can accomplish include:

\begin{itemize}
    \item \textbf{Water scene filming:} WaveShot can capture stationary objects such as anchored boats, water buoys, or waterfront landscapes in natural water conditions. It is used to capture details of static objects such as textures of surrounding plants, floating objects on the water surface, or rocks by the water's edge. Additionally, WaveShot can identify and track rapidly moving objects such as kayaks, water scooters, or swimmers from a distance and follow the targets to obtain clear dynamic shots.
    \item \textbf{Monocular depth estimation:} By processing images captured by the portable waterborne device, depth videos are generated through the Depth Anything \cite{yang2024depth} project. This includes not only visual information of regular videos but also adds depth information to each pixel, providing additional dimensions and insights for various applications. It provides crucial spatial information to help identify potential obstacles and assess the navigability of rescue areas.
\end{itemize}

\begin{figure}[t]
\centerline{\includegraphics[width=0.5\textwidth]{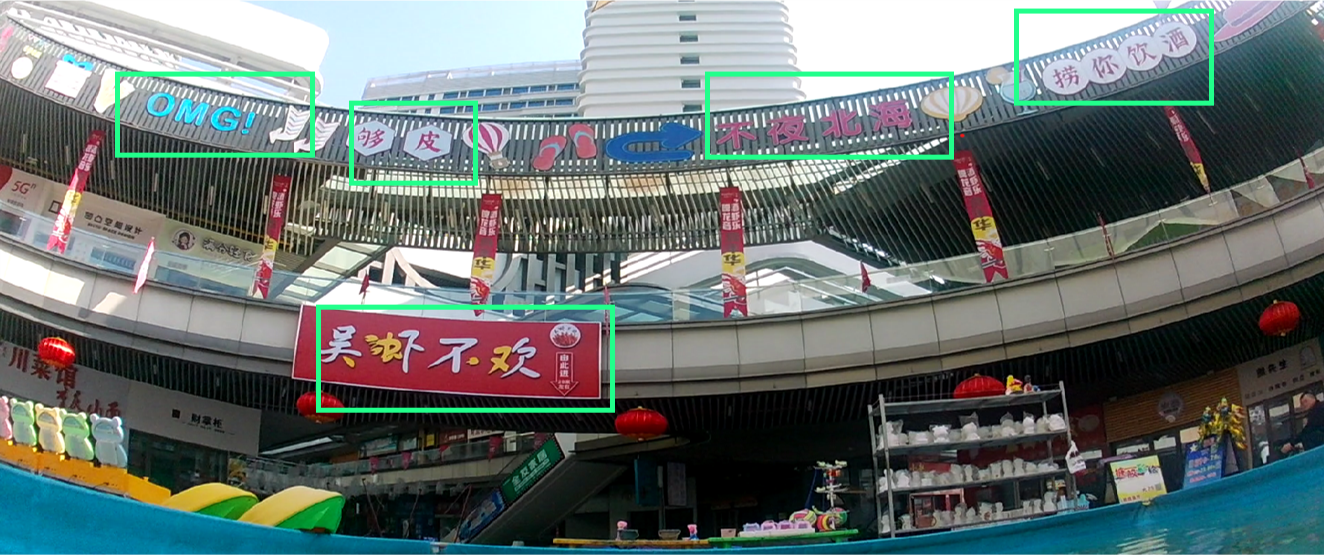}}
\caption{Static Object Shooting Task: The scene is set in a shopping mall.}
\label{fig:text}
\end{figure}

\section{TASKS}

This study is designed to evaluate the capabilities of the WaveShot, a portable aquatic drone, in capturing video footage of both static and moving objects on water. The tasks are divided into three main categories: static object shooting, moving object shooting and depth estimation. Each category aims to test different aspects of the WaveShot's performance, including stability, clarity, tracking ability, and operational flexibility under various environmental conditions.

\subsection{Static Object Shooting}

For \textbf{\textit{Close-range Detail Shooting Task}}, the task focuses on the WaveShot's ability to capture detailed video footage of a static object from a close range. It tests the drone's precision in controlling its distance and angle relative to the object, particularly when capturing specific parts of the object, like surface textures or color gradients. The task evaluates the drone's maneuverability and precision in capturing high-quality detailed shots.

\subsection{Moving Object Shooting}

For \textbf{\textit{Moving Water Object from Distance to Close Range Shooting Task}}, the WaveShot must identify a moving target on the water from a distance and follow it as it moves closer, ultimately achieving clear, close-range footage. This task tests the drone's target recognition and dynamic tracking capabilities, as well as its ability to remain stable during the approach. The challenge lies in capturing the object's details and maintaining shooting continuity as the distance and possibly the speed of the object change.

For \textbf{\textit{Moving Water Object from Close Range to Distance Shooting Task}}, 
this task requires the WaveShot to start capturing footage of a moving target on the water from a close range and continue to maintain the target within the frame as it moves to a farther distance. This task similarly tests the drone's tracking and shooting stability, especially in adjusting the shooting angle to keep the moving target in a clear and central position as it moves away. The challenge is to maintain effective tracking of a moving object at a distance while ensuring the stability of the footage.

\subsection{Depth Estimation}

For \textbf{\textit{Water Scene Depth Estimation Task}}, It involves Depth Estimation applying the Depth Anything \cite{yang2024depth} model to accurately estimate the depth of various water scenes with a particular focus on images containing fishing boats and unmanned boats. The process is encapsulated within several pivotal steps:

\begin{enumerate}
    \item \textbf{Data Preprocessing}: Input images \(I\) undergo preprocessing to fit the depth estimation model's requirements, including scaling, cropping, and normalization.
    
    \item \textbf{Forward Propagation in Depth Estimation Network}: The model \(S\) performs forward propagation on the preprocessed image \(I'\), yielding the predicted depth map \(D'\) as \(D' = S(I'; \theta)\), where \( \theta \) symbolizes the model parameters.
    
    \item \textbf{Loss Function Calculation}: 

    Loss Function for Labeled Dataset (Affine-Invariant Mean Absolute Error):
  
      \[
      \mathcal{L}_{\text{labeled}} = \frac{1}{W \times H} \sum_{i=1}^{W} \sum_{j=1}^{H} \left| \hat{d}_{i,j} - d_{i,j} \right|
      \]
      
      Here, $\hat{d}_{i,j}$ represents the predicted depth value for pixel $(i,j)$, $d_{i,j}$ represents the true depth value for pixel $(i,j)$, and $W$ and $H$ denote the width and height of the image, respectively.

    Loss Function for Unlabeled Dataset (including pseudo labels and CutMix parts):
  
      \[
      \mathcal{L}_{\text{unlabeled}} = \mathcal{L}_{\text{pseudo}} + \mathcal{L}_{\text{cutmix}}
      \]
      
      Here, $\mathcal{L}_{\text{pseudo}}$ denotes the loss based on pseudo labels generated by the teacher model, and $\mathcal{L}_{\text{cutmix}}$ denotes the loss when generating mixed images using the CutMix data augmentation technique.

    Feature Alignment Loss:
  
      \[
      \mathcal{L}_{\text{align}} = \frac{1}{N} \sum_{i=1}^{N} \max(0, \alpha - \cos(\mathbf{f}_i, \mathbf{f}_i^{\text{pre}}))
      \]
      
      Here, $\mathbf{f}_i$ represents the $i$th feature vector, $\mathbf{f}_i^{\text{pre}}$ represents the corresponding feature vector in the pretrained model, $\cos(\cdot, \cdot)$ denotes the cosine similarity, $\alpha$ denotes the similarity threshold, and $N$ denotes the total number of feature vectors.

    Combining the above, the complete loss function can be expressed as:
    
    \[
    \mathcal{L}_{\text{total}} = \mathcal{L}_{\text{labeled}} + \mathcal{L}_{\text{unlabeled}} + \lambda \mathcal{L}_{\text{align}}
    \]
    
    Where, $\lambda$ denotes the weighting coefficient for the feature alignment loss.

    \item \textbf{Optimization and Model Update}: The AdamW optimizer refines model parameters \( \theta \) to minimize the loss \( L \), incorporating a linear learning rate decay.
    
    \item \textbf{Post-Processing}: The predicted depth map \(D'\) undergoes post-processing, such as scale and offset adjustments, to align depth values with real-world measurements.
\end{enumerate}

\begin{figure*}[htbp]
\centering
\includegraphics[width=\textwidth,trim={0.3cm 0cm 0cm 0cm},clip]{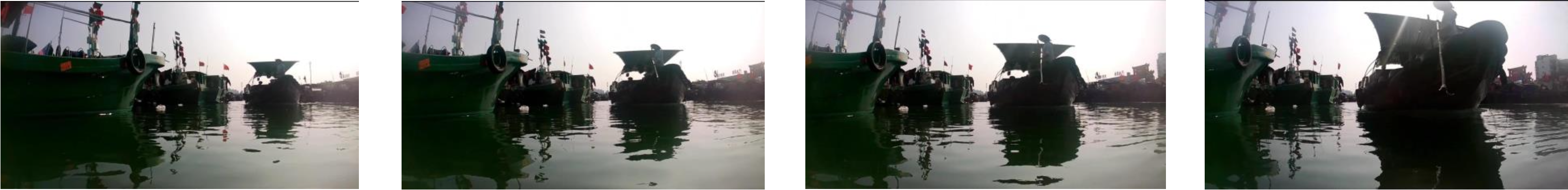}\\
\raggedright\small\textbf{\textit{Fishing Boat:}} WaveShot captures a fishing boat moving from a distance to a closer position in low-light conditions. \\[1ex]

\includegraphics[width=\textwidth,trim={0.3cm 0cm 0cm 0cm},clip]{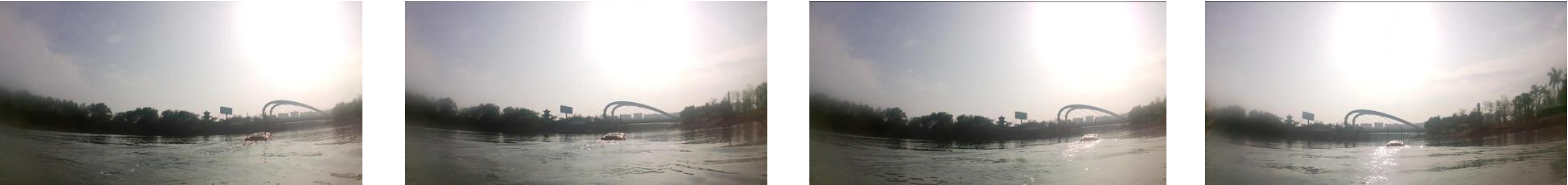}\\
\raggedright\small\textbf{\textit{Unmanned Boat:}} WaveShot captures an unmanned boat moving from a closer position to a distance under direct sunlight. \\[1ex]

\caption{\small \textbf{\textit{Task of Moving Object Capture on Water}}.}
\label{fig:moving_object}
\end{figure*}

\begin{figure*}[htbp]
\centering
\includegraphics[width=\textwidth,trim={0.3cm 0cm 0cm 0cm},clip]{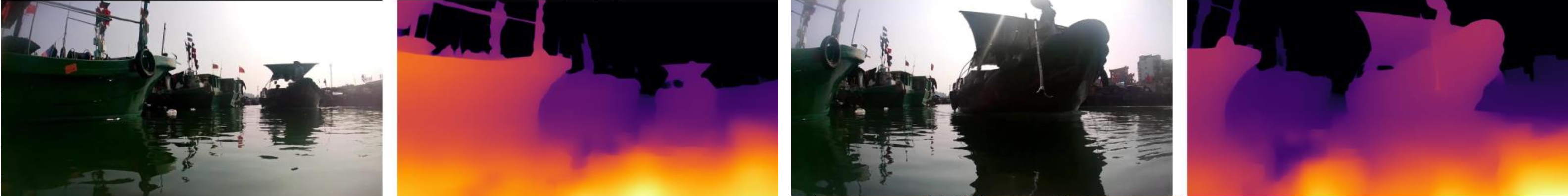}\\
\raggedright\small\textbf{\textit{Left:}} long-range fishing boats and their corresponding depth maps. \textbf{\textit{Right:}} short-range fishing boats and their corresponding depth maps.\\[1ex]

\includegraphics[width=\textwidth,trim={0.3cm 0cm 0cm 0cm},clip]{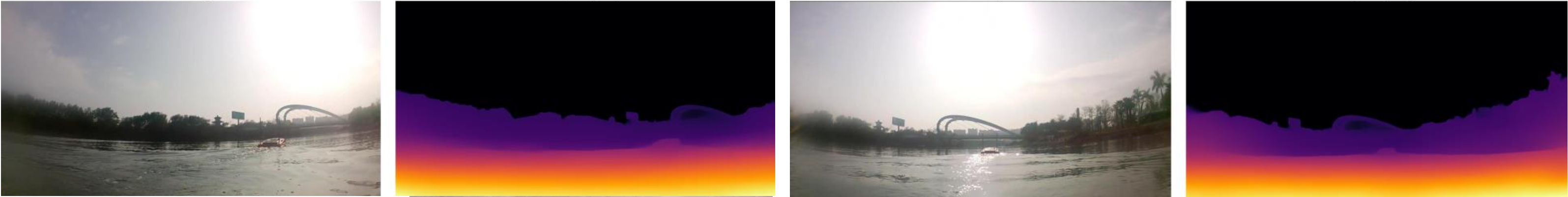}\\
\raggedright\small\textbf{\textit{Left:}} a short-range unmanned boat and its depth map. \textbf{\textit{Right:}} a long-range unmanned boat and its depth map.\\[1ex]

\caption{\small \textbf{\textit{Task of Water Scene Depth Estimation}}.}
\label{fig:depth}
\end{figure*}

\begin{table}[t]
	\centering
	\caption{Text clarity evaluation}
	\vspace{-0.2cm}
	\renewcommand\arraystretch{1.5}
	\renewcommand\tabcolsep{4pt}
	\scriptsize
	\begin{tabular}{{p{7cm}l}}
	\toprule
	\textbf{Text} & \textbf{Clarity} \\
  	\midrule[0.03cm]
	
	"OMG!" & 10 \\
	\midrule[0.01cm]

	Text on the middle-sized banner in the center & 8 \\
	\midrule[0.01cm]

	Text on the small white banner in the top left & 7 \\
	\midrule[0.01cm]

	Text on the small white banner in the top right & 7 \\
	\midrule[0.01cm]

	Text on the bottom red banner & 9 \\
	\midrule[0.01cm]

	\textbf{Average} & \textbf{8} \\
  	\bottomrule[0.03cm]
	\end{tabular}
	\label{tab:text_clarity}
	\vspace{-0.5cm}
\end{table}

\section{EXPERIMENTS}

In this study, we utilized the advanced capabilities of OpenAI's GPT-4V model \cite{openai2024gpt4} to evaluate the performance of WaveShot across various tasks. The GPT-4V model is designed to respond to queries based on given text and images, which allows for a comprehensive assessment of WaveShot's capabilities through a series of designed questions and corresponding image sequences. This approach aims to provide a nuanced understanding of WaveShot's performance in capturing video details in water-based environments.

\subsection{Static Object Detail Capture in Close Proximity}
For the task, The first experimental setup involved a 15-meter-long and 6-meter-wide outdoor pool located at a shopping center to evaluate WaveShot's ability to capture static objects in detail at close range. The pool scene, surrounded by shops displaying various textual advertisements, served as an ideal setting for assessing the clarity with which WaveShot could capture text in images in Figure~\ref{fig:text}. With an overall text clarity score of 8, the results indicated that WaveShot could capture static object details with high clarity under near-distance and favorable lighting conditions in Table~\ref{tab:text_clarity}. This suggests that WaveShot is capable of meeting general photography needs, especially in static object scenarios.

\subsection{Moving Object Capture on Water}
For the experimental shooting of moving objects on water, we selected two objects to capture: an unmanned boat and a fishing boat in Figure~\ref{fig:moving_object}. When it came to filming the fishing boat up close, WaveShot was positioned slightly lower. Its close-up lens captured some details of the boat, such as the color of the hull, its structure, and items on board. The main parts of the boat, such as the bow and the sides, remained in focus and clear. However, due to the gloomy weather or low lighting conditions during the filming, the brightness and contrast of the content were affected. The video displayed WaveShot's capabilities in both stationary and dynamic shots, particularly in maintaining stable filming. 

In another experiment, WaveShot captured the process of a small unmanned boat moving away from the camera. Each frame captured it at a further distance. During the filming, WaveShot appeared to successfully track the moving unmanned boat and kept it centered in the frame as it moved away. This was beneficial in maintaining focus on the subject. However, the camera was facing a strong light source, possibly the sun, which resulted in overexposed images. This caused the loss of detail in the sky and surrounding scenery. In the initial frames, the boat remained relatively clear, but as the distance increased, the clarity decreased, especially when filming a moving subject. Overall, WaveShot performed well in terms of stability and subject tracking. However, it faced challenges in exposure control and clarity, particularly in long-distance shots and under varying lighting conditions.

\subsection{Water Scene Depth Estimation}
Using video material from the moving object capture experiments, relative depth estimation for a fishing boat scene and an unmanned boat scene was conducted in Figure~\ref{fig:depth}. The depth estimation utilized color coding to indicate the distance of objects from the observer, with warm colors for closer objects and cool colors for distant objects. While the depth estimation maps generally reflected the depth information accurately, some inaccuracies were noted at edges and transition areas. The depth maps showed effective distance estimation between the unmanned boat and its surroundings, although finer details like the antennas on the unmanned boat were not captured, indicating limitations in detail retention in multi-object water scenes.

Overall, the experiments conducted provide valuable insights into WaveShot's capabilities in capturing details and estimating depths in water-based environments, highlighting its strengths and areas for improvement in both static and dynamic conditions.

\section{CONCLUSIONS}

In this work, we introduced WaveShot, a novel compact and portable unmanned surface vehicle designed for filming and media production in aquatic environments. By integrating differential propulsion, onboard sensing and processing, remote operation, and action camera functionality, WaveShot enables smooth tracking shots and stable image capture on the water surface. It provides a flexible and accessible platform for aquatic photography and extends the toolkit for film and media production. This research demonstrates the potential of unmanned surface vehicles in creative arts and other emerging applications. Addressing current limitations and advancing the concept of WaveShot can lead to more dynamic image capture and novel visual perspectives in aquatic environments.

Our experimental results demonstrate the effectiveness of WaveShot in filming both static and dynamic objects on water. The intuitive control interface, allowing even beginners to quickly learn complex operations. Qualitative and quantitative analyses suggest that WaveShot has advantages in stability, detail capture, tracking, and depth estimation using state-of-the-art monocular models.

However, the current WaveShot prototype and methods have some limitations. The wireless transmission range is currently limited to approximately 1 kilometer, restricting the operational radius. Additionally, factors such as weather, lighting, and water flow turbulence may negatively impact image quality and model performance. Occlusions, reflections, and the absorbance properties of water still pose challenges for accurate depth estimation, especially over longer distances.

This work opens up many promising directions for future development. We hope to integrate waterproof gimbal stabilization systems in future work to further enhance filming stability in adverse conditions. On the software side, the application of more powerful control and computer vision techniques tailored for aquatic filming can optimize the filming process in water scenes.

\addtolength{\textheight}{-12cm}   




\bibliographystyle{IEEEtran}
\bibliography{main}

\end{document}